\documentclass[letterpaper, 10 pt, conference]{IEEEConference}
\IEEEoverridecommandlockouts
\usepackage{bm}
\usepackage{svg}
\usepackage{times}
\usepackage{epsfig}
\usepackage{amsmath}
\usepackage{amssymb}
\usepackage{graphics}
\usepackage{multirow}
\usepackage{booktabs}
\usepackage{algorithm}
\usepackage{subcaption}
\usepackage[noend]{algpseudocode}

\let\labelindent\relax
\usepackage{enumitem}

\graphicspath{{./figures/svg_to_pdf/}}

\DeclareMathOperator*{\maxdist}{d_{\text{max}}}
\DeclareMathOperator*{\maxcost}{c_{\text{max}}}

\title{\LARGE \bf
Safe Multi-Agent Navigation guided by Goal-Conditioned Safe Reinforcement Learning
}

\author{Meng Feng$^{*1}$, Viraj Parimi$^{*1}$ and Brian Williams $^{1}$
\thanks{$^{1}$ Computer Science and Artificial Intelligence Laboratory, Massachusetts Institute of Technology, Cambridge, MA 01239. Corresponding at {\tt\footnotesize \{mfeng,vparimi,williams\}@mit.edu}. *These authors contributed equally to the paper.}
\thanks{This work was also supported by Defence Science and Technology Agency, Singapore}
}

\begin{document}

\maketitle
\thispagestyle{empty}
\pagestyle{empty}

\begin{abstract}
Safe navigation is essential for autonomous systems operating in hazardous environments. Traditional planning methods are effective for solving long-horizon tasks but depend on the availability of a graph representation with predefined distance metrics. In contrast, safe Reinforcement Learning (RL) is capable of learning complex behaviors without relying on manual heuristics but fails to solve long-horizon tasks, particularly in goal-conditioned and multi-agent scenarios.

In this paper, we introduce a novel method that integrates the strengths of both planning and safe RL. Our method leverages goal-conditioned RL (GCRL) and safe RL to learn a goal-conditioned policy for navigation while concurrently estimating cumulative distance and safety levels using learned value functions via an automated self-training algorithm. By constructing a graph with states from the replay buffer, our method prunes unsafe edges and generates a waypoint-based plan that the agent then executes by following those waypoints sequentially until their goal locations are reached. This graph pruning and planning approach via the learned value functions allows our approach to flexibly balance the trade-off between faster and safer routes especially over extended horizons.

Utilizing this unified high-level graph and a shared low-level safe GCRL policy, we extend this approach to address the multi-agent safe navigation problem. In particular, we leverage Conflict-Based Search (CBS) to create waypoint-based plans for multiple agents allowing for their safer navigation over extended horizons. This integration enhances the scalability of goal-conditioned safe RL in multi-agent scenarios, enabling efficient coordination among agents. Extensive benchmarking against state-of-the-art baselines demonstrates the effectiveness of our method in achieving distance goals safely for multiple agents in complex and hazardous environments. Our code and further details about or work is available at https://safe-visual-mapf-mers.csail.mit.edu/.
\end{abstract}

\begin{figure*}[t]
    \centering  
    \includegraphics[width=\textwidth]{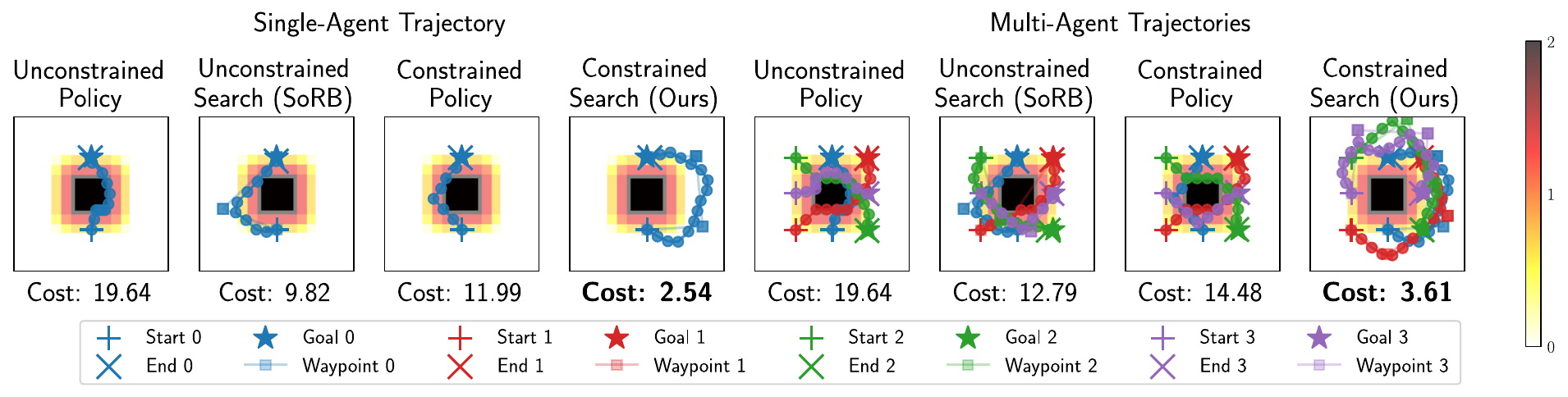}
    \caption{Comparison of different approaches on the 2D navigation problem for a single agent (left) and four agents (right). Baselines tend to traverse high-cost regions, leading to suboptimal paths. In contrast, our approach finds a longer but lower-cost path by avoiding these regions. For both unconstrained and constrained policies, collision avoidance in the multi-agent case is ensured by having agents wait when another agent is nearby.}
    \label{fig:pointenv_illustration_together}
\end{figure*}

\section{Introduction}
In many real-world scenarios like search-and-rescue or environmental monitoring, autonomous agents must navigate to distant locations safely as repairs can be costly and time-consuming. This challenge becomes even more critical in multi-agent settings, such as drone swarms or robotic teams, where all agents must coordinate to reach their goals while managing risks throughout the entire mission. Ensuring safety in such complex environments is paramount, as failure to do so could compromise the entire operation. Recent advancements in Multi-Agent Path Finding (MAPF) \cite{mapf_benchmarks, yu2013structure, infomapf} demonstrate the ability of planners to efficiently generate path plans for multiple agents by decomposing a multi-agent problem into a series of single-agent problems. However, these approaches assume access to a structured graph representation of the environment, consisting of valid nodes and edges, annotated with utility information like safety levels and estimated travel times, limiting their applicability in scenarios where agents only have access to high-level observations, such as images.

In contrast, learning-based approaches such as deep reinforcement learning (RL) excel at learning state utilities from high-dimensional observations such as images. In particular, goal-conditioned RL (GCRL) \cite{mirowski2017learning, pmlr-v37-schaul15,pong2020temporaldifferencemodelsmodelfree} has been effective in enabling agents to achieve goals of interest in various domains, including navigation and manipulation. However, learning policies for distant goals remains challenging, as performance often degrades as the task horizon increases \cite{levy2019hierarchical,nachum2018dataefficient} and requires significant reward shaping \cite{chiang2019learning} or expert demonstrations \cite{lynch2019learning,nair2018overcoming}, which can limit the policy's ability to discover novel solutions. The challenge is compounded when agents must remain safe while navigating over long horizons. While constrained RL \cite{altman2021constrained} has demonstrated the potential to incorporate safety constraints into policy learning, reaching a diverse set of distant goals under such constraints, particularly in multi-agent contexts, remains as an open research problem.

Prior research has tackled long-horizon problems using GCRL combined with graph search techniques \cite{eysenbach2019search}. By constructing an intermediate graph representation of the environment from the replay buffer and using search-based methods to generate a waypoint-based plan, these methods decompose tasks into a series of shorter and easier goal-reaching tasks. However, existing approaches are limited to single-agent scenarios without considering any safety aspects. This paper extends these methods to address multi-agent, long-horizon, goal-conditioned tasks ensuring that the agents reach their goals safely.

Our approach leverages GCRL and safe RL via a novel self-training algorithm to learn distance and safety predictors which allows us to dynamically construct an intermediate graph from the replay buffer. This enables greater planning flexibility that balances the trade-off between faster and safer routes depending on user-preference. By utilizing this intermediate graph representation, we extend the range of problems that MAPF approaches can solve, especially in challenging, image-based environments, effectively bridging the gap between planning and learning in multi-agent settings. Additionally, we empirically show that our approach can plan a series of safer sub-goals for multiple agents without requiring new training for each additional agent, thereby achieving scalability. In summary, we present a scalable and safer approach for navigating multiple agents to their distance goals in sparse-reward environments, broadening the scope of MAPF through the integration of safe GCRL.

To give an overview of the rest of the paper, first in Section \ref{preliminaries} we define the relevant preliminary information to set the stage for our approach. In Section \ref{approach}, we describe our solution.  Finally, Section \ref{experiments} presents our experimental results, where we compare our algorithm against state-of-the-art baselines.  

\section{Preliminaries}
\label{preliminaries}

\textbf{Goal-Conditioned Reinforcement Learning}. In goal-conditioned RL (GCRL), an agent interacts with an environment modeled as an augmented Markov Decision Process (MDP) \((\mathcal{S}, \mathcal{A}, P, R, \gamma)\), where \(\mathcal{S}\) is the state space, \(\mathcal{A}\) is the action space, \(P(s_{t+1}|s_t, a_t, s_g)\) is the transition probability, \(R(s, a, s_g)\) is the reward function, and \(\gamma \in [0, 1)\) is the discount factor. Unlike vanilla MDP, GCRL includes a goal $s_g \in \mathcal{S}$ in the problem formulation. The agent’s objective is to learn a policy \(\pi(a|s, s_g)\) that maximizes the expected cumulative reward, conditioned on both the current state \(s\) and the goal \(s_g\). The reward function \(R(s, a, s_g)\) is goal-dependent, reflecting how well the actions align with the desired goal. In practice, GCRL excels at learning short-horizon skills often with the help of reward shaping \cite{chiang2019learning} and human demonstrations \cite{lynch2019learning,nair2018overcoming} to enhance performance.

\textbf{Bridge Planning and Single-Agent RL}. To address long-horizon problems, \cite{eysenbach2019search} proposed Search on Replay Buffer (SoRB) that effectively integrates RL with single-agent planning. Specifically, SoRB defines $R(s, a, s_g) = -1$ and $\gamma = 1$ for the target environment. It then trains a goal-conditioned policy \(\pi(a \mid s, s_g)\) and the associated value function $V(s, s_g)$ using standard off-policy RL algorithms \cite{lillicrap2019continuouscontroldeepreinforcement}. Assuming that \(\pi(a \mid s, s_g)\) is optimal, the value function $V(s, s_g)$ can be used to approximate the distance between a state $s$ and a goal state $s_g$:
\begin{equation}
    V(s,s_{g})=-d_{\textnormal{sp}}\left(s,s_{g}\right)\label{eq:distance_from_value_function}
\end{equation}
where $d_{\textnormal{sp}}$ is the shortest path distance from $s$ to $s_{g}$, or equivalently, the expected number of steps to reach $s_{g}$ from $s$ under the optimal policy $\pi$.

SoRB then constructs a weighted, directed graph $\mathcal{G}$ from states that are either drawn from the replay buffer or randomly sampled form the environment state space. Each node in the graph corresponds to an observation and edges are added based on the predicted distance $V(s_1, s_2)$ between the node pairs $(s_1, s_2)$. Node pairs $V(s_1, s_2) > \maxdist$ are excluded as they are considered potentially unreachable. 

Finally, SoRB generates a waypoint-based plan by searching on this graph $\mathcal{G}$. The agent sequentially navigates to the individual waypoints before heading towards the final goal. This process is executed in either an open-loop or closed-loop manner. Notably, SoRB employs a greedy strategy and does not take the safety of the agent into account. However, when safety is factored in, longer and safer paths may be preferred over shorter yet riskier paths. 

\textbf{Constrained Reinforcement Learning \label{subsec:crl}}. To capture the concept of risk and safety, we build our learning algorithm within the framework of Constrained RL (CRL). CRL solves a constrained Markov Decision Process (CMDP) \cite{altman2021constrained} problem that augments MDP with extra constraints that restrict the set of allowable policies. CMDP defines an \textit{auxiliary cost} function $C:S\times A\times S\rightarrow\mathbb{R}^+$ and a cost limit $\delta \in \mathbb{R}^+$. For conciseness, we will refer to auxiliary cost as \textit{cost}. The cost function serves as a generic measure of safety. For example, regions of high environmental uncertainty may be deemed unsafe and induce higher costs. In this work, we do not require the cost function to carry any physical interpretation (e.g., uncertainty). Let $J_{C}\left(\pi\right)=\underset{\tau\sim\pi}{\mathbb{E}}\left[\sum_{t=0}^{\infty}\gamma^{t}C\left(s_{t},a_{t},s_{t+1}\right)\right]$ denote the expected discounted \textit{cost return} following policy $\pi$. The set of feasible stationary policies for a CMDP is then
\begin{equation}
    \Pi_{C}\doteq\left\{ \pi\in\Pi:J_{C}\left(\pi\right)\le d\right\} 
\end{equation}
Standard off-policy methods can be adapted to solve CRL problems by including Lagrange multipliers to penalize unsafe actions that lead to the violation of cost constraints \cite{Ray2019}. It is crucial to note that, in CRL, the cost function and cost limit is task-dependent and is not applicable to the goal-conditioned setting where each goal is akin to a distinct task. For a more detailed reference on CRL, we refer our readers to \cite{schulman2015trust,schulman2017proximal,sutton2018reinforcement}. , 

\textbf{Multi-Agent Path Finding (MAPF)}. Focusing on multi-agent settings, we explore the planning through the lens of a Multi-Agent Path Finding (MAPF) problem \cite{mapf_benchmarks, yu2013structure}. A MAPF problem involves a weighted graph $\mathcal{G}(\mathcal{V}, \mathcal{E})$ and a set of agents $A = {a_1, a_2, \ldots, a_k}$, each with a unique start $s_i \in \mathcal{V}$ and a goal $g_i \in \mathcal{V}$. Agents can either wait or move to adjacent vertices along edges in $\mathcal{E}$, with each action incurring a penalty defined by $\mathcal{G}$. Once an agent reaches its goal, no further penalty is incurred.

\begin{figure*}[ht]
    \centering  
    \includegraphics[width=0.85\textwidth]{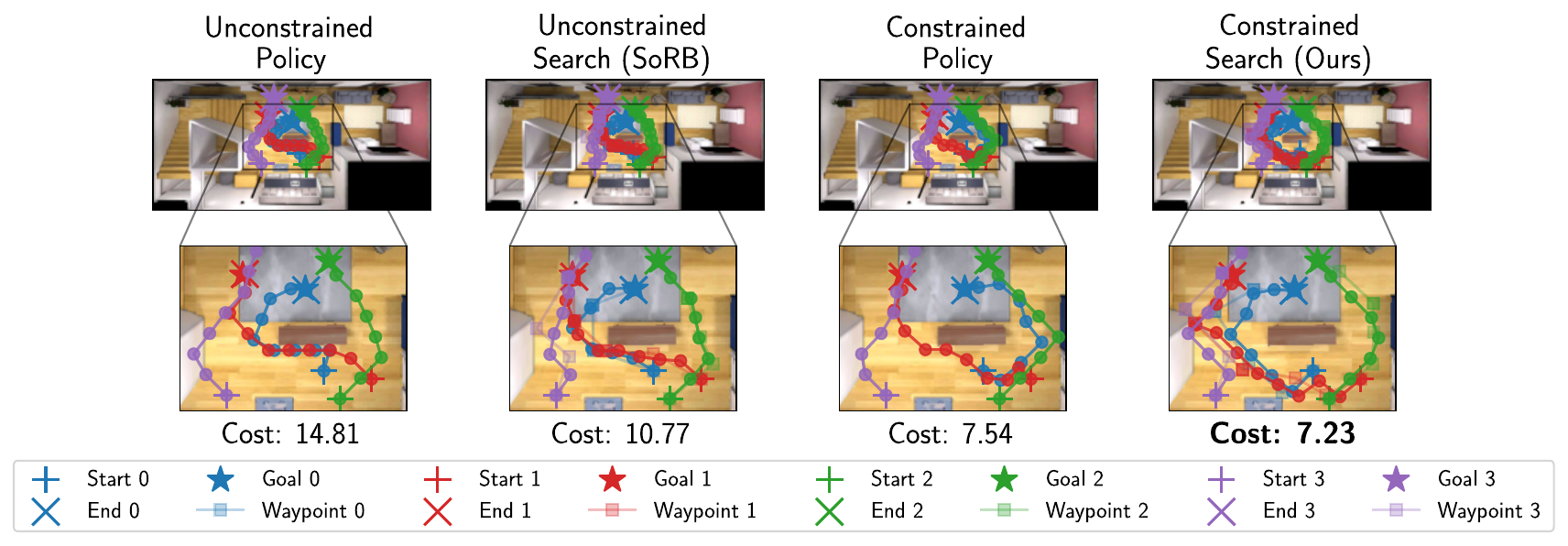}
    \caption{Comparison of different approaches on the SC3 Staging 11 map for four agents shows that unconstrained baselines fail to avoid the furniture, whereas the constrained policy and our approach successfully navigates around it. Our approach goes further by re-routing the agents away from the furniture, ensuring the safest behavior. Note that for both unconstrained and constrained policies, inter-agent collision avoidance is ensured by having agents wait when another agent is nearby.}
    \label{fig:habitatenv_multi_illustration}
\end{figure*}

\section{Approach}
\label{approach}

\textbf{Goal-Conditioned Safe Reinforcement Learning}. Inspired by SoRB \cite{eysenbach2019search}, we would like to train an agent that can predict both distance and cost among state pairs. Meanwhile, the agent should also be capable of safely navigating amongst them. Since the value functions are conditioned on the associated policy, in order to obtain the distance and cost predictions following the shortest path, it is critical to first train an unconstrained agent that includes a goal-conditioned policy $\pi(s, a, s_g)$, a $Q$-function for cumulative reward $Q(s, a, s_g)$, and another $Q$-function for cumulative cost $Q_C(s, a, s_g)$. For the agent to learn safer navigation, we then fine-tune the unconstrained policy $\pi$ by applying constrained RL methods which outputs a goal-conditioned safe policy $\pi_{c}(s, a, s_g)$ and its associated value functions. As previously noted, the cost limit from CRL is not directly applicable to the goal-conditioned setting. Therefore, we redefine the cost limit as a hyper-parameter, a soft constraint that penalizes unsafe actions without requiring strict compliance. However, we note that when the choice of cost limit is too low (penalty is too high), the resultant policy would focus too much on reducing costs rather than reaching the goals. In this work, we use the Lagrangian actor critic \cite{Ray2019} thanks to its simplicity.

\textbf{Self-Sampling and Training}. To train a goal-conditioned agent, a set of goal states need to be sampled for the agent to explore and practice reaching them. The design of sampling strategy is crucial for successful training of a goal-conditioned agent. SoRB \cite{eysenbach2019search} accomplishes this by prioritizing nearby goal samples to improve the success rate of short-horizon navigation. However, it relies on an oracle with access to the ground-truth positions rather than the raw observations (e.g., images) to sample goals of desirable distances. This practice is unrealistic as such oracles often do not exist. Instead, we propose Algorithm \ref{alg:training}, a self-sampling and training algorithm which, in essence, uses the agent's $Q$-functions to evaluate and select training samples with a diverse set of distances and cost targets. As the $Q$-functions become more accurate, the quality of the samples improve, which then generates better training samples to improve the policy.
\begin{algorithm}
    \caption{Self-Sampling and Training of Goal-Conditioned Actor Critic}
    \label{alg:training}
    \textbf{Inputs}: 
    \begin{tabular}[t]{ll}
        Environment $\mathrm{E}$, Goal-Conditioned Policy $\pi$, \\
        Associated $Q$-functions for rewards or auxiliary \\ 
        costs $Q(s, a, s_g)$, Desirable sample target for \\ 
        rewards or costs $\tau$, Population size $N$, \\
        Number of training problems per batch $K$
    \end{tabular}
    \textbf{Outputs}: Updated $\pi(s,a,s_g)$, $Q(s, a, s_g)$
    \begin{algorithmic}[1]
    \State $\mathcal{P} \gets$ Initialize an empty training set that will contain start-goal pairs spaced at the desired targets.
    \For{each batch}
        \State $\{(s_i,s_j)\}_N \gets$ Randomly sample state pairs from $\mathrm{E}$
        \State $\{v_{ij}\}_N \gets Q(s_i, \pi(s_i, a, s_j), s_j)$
        \State $\{l_{ij}\}_N \gets $ $L^2$ distance to $\tau$
        \State Find $\{(s_i,s_j)\}$ corresponding to lowest $K$  $\{l_{ij}\}_N$
        \State Add the selected $\{(s_i,s_j)\}$ to the training set
        \State Train the agent with $\mathcal{P}$ until $\mathcal{P}$ is depleted
    \EndFor
    \end{algorithmic}
\end{algorithm}

\textbf{Categorical Cost Estimate}. Training a robust goal-conditioned value function is challenging yet crucial for planning efficient waypoints \cite{eysenbach2019search}. Incorporating distributional RL \cite{bellemare2017distributional} has shown to be an effective strategy in improving the quality of goal-conditioned value functions thereby enhancing the planning process. Therefore, we follow \cite{bellemare2017distributional} to model the goal-conditioned value function for costs as a categorical distribution $Q_C^{\pi} \sim \textnormal{Cat}(N,p_{\pi}(z\mid s,a,g))$ within a range of $[V_{\min}^{c},V_{\max}^{c}],V_{\min}^{c},V_{\max}^{c}\in\mathbb{R}^{+}$, where $N \in \mathbb{N}$ is the number of evenly discrete classes, $p_{\pi}(z\mid s,a,g)$ is the likelihood of each specific cost outcome predicted by the cost value function conditioned on the current policy $\pi$. The class supports can be described by a set of atoms $\{z_{i} = V_{\min}^{c} + i \Delta z: 0 \le i <N \}, \Delta z \triangleq (V_{\max}^{c} - V_{\min}^{c}) / (N-1)$. Consequently, the expected value of the distribution is used estimate the goal-conditioned cost:
\begin{equation}
    c_{\pi}(s_{1}, g)=\underset{p(z \mid s, a, g)}{\mathbb{E}} \left[ z \right]
    \label{eq:cat_cost}
\end{equation}
Since $Q_C^{\pi}$ is a categorical distribution, standard Bellman update is not applicable. Instead, we follow the Categorical Algorithm \cite{bellemare2017distributional} to update $Q_C(s, a, s_g)$. 

\textbf{Graph Construction from Replay Buffer}. With a trained agent, we use the observations (e.g., images) in the replay buffer $\mathcal{B}$ to construct a weighted, directed graph $\mathcal{G}$. Each edge has a predicted distance and cost for the state pair $(s_i, s_j)$:
\begin{align}
    d_{sp} \approx d_{\pi} \gets & \; Q(s,\pi(s_{i},a,s_{g}),s_{j})
    \label{eq:predict_dist} \\
    c_{sp} \approx c_{\pi} \gets & \; Q_{C}(s,\pi(s_{i},a,s_{g}),s_{j})
    \label{eq:predict_cost}
\end{align}
\begin{align*} 
    & \mathcal{G} \triangleq \left( \mathcal{V}, \mathcal{E}, \mathcal{W}_{d}, \mathcal{W}_{c} \right) \\
    \textnormal{where } & \mathcal{V} = \mathcal{B} \\
    & \mathcal{E} = \mathcal{B} \times \mathcal{B} = \left\{ e_{s_{i}\rightarrow s_{j}} \vert s_{i}, s_{j} \in \mathcal{B} \right\} \\
    & \mathcal{W}_{d} \left( e_{s_{i} \rightarrow s_{j}} \right) = 
        \begin{cases}
            d_{\pi}(s_{i}, s_{j}) & \textnormal{if }d_{\pi}(s_{i}, s_{j}) < \maxdist \\
            \infty & \textnormal{otherwise}
        \end{cases} \\
    & \mathcal{W}_{c} \left( e_{s_{i} \rightarrow s_{j}} \right) = 
        \begin{cases}
            c_{\pi}(s_{i}, s_{j}) & \textnormal{if }c_{\pi}(s_{i}, s_{j}) < \maxcost \\
            \infty & \textnormal{otherwise}
        \end{cases}
\end{align*}
We exclude edges whose distance exceeds a maximum cutoff, $\maxdist$, or whose cost surpasses a maximum cutoff, $\maxcost$. Both $\maxdist$ and $\maxcost$ are treated as hyper-parameters.
\textbf{Conflict-Based Search (CBS) for Multi-Agent Planning} \cite{sharon-cbs} is a widely used MAPF algorithm that addresses the problem's exponential state space by decomposing it into a series of single-agent path planning problems which are then combined using efficient conflict-resolution strategies. CBS operates on two levels: the high-level search, which manages different constraints on agents' path plans, and the low-level search, where a space-time A* algorithm is used to find paths for each agent independently while satisfying the constraints imposed by the high-level search tree.

In CBS, constraints are typically defined in the form $\langle a_i, v, t \rangle$, indicating that agent $a_i$ is prohibited from visiting vertex $v \in \mathcal{V}$ at time step $t$. The high level search constructs a \textit{constraint tree} starting from a root node with no constraints. The tree expands in a best-first manner where each successor node inherits its parent's constraints and adds a new constraint for a single agent. The goal of CBS is to reach a node in the constraint tree where all paths of all agents are conflict-free. This two-level structure of CBS effectively manages the complexity of MAPF problems by isolating single-agent planning tasks and resolving conflicts incrementally.

\textbf{High-Level Overview}. In summary, our approach leverages goal-conditioned safe RL augmented with self-sampling and training algorithm \ref{alg:training} to build a graph from the replay buffer annotated with predicted distances and cost estimates using the $Q$-functions $Q^{\pi}, Q^{\pi}_C$ of the \textit{unconstrained} policy $\pi$. CBS then utilizes this intermediate graph to generate effective and safer sequential waypoints for the agents, which are followed using the trained low-level \textit{safe} GCRL policy $\pi_c$ for enhanced bi-level safety.

\begin{figure}[ht]
    \centering
    \begin{subfigure}[b]{0.35\textwidth}
       \includegraphics[width=\linewidth]{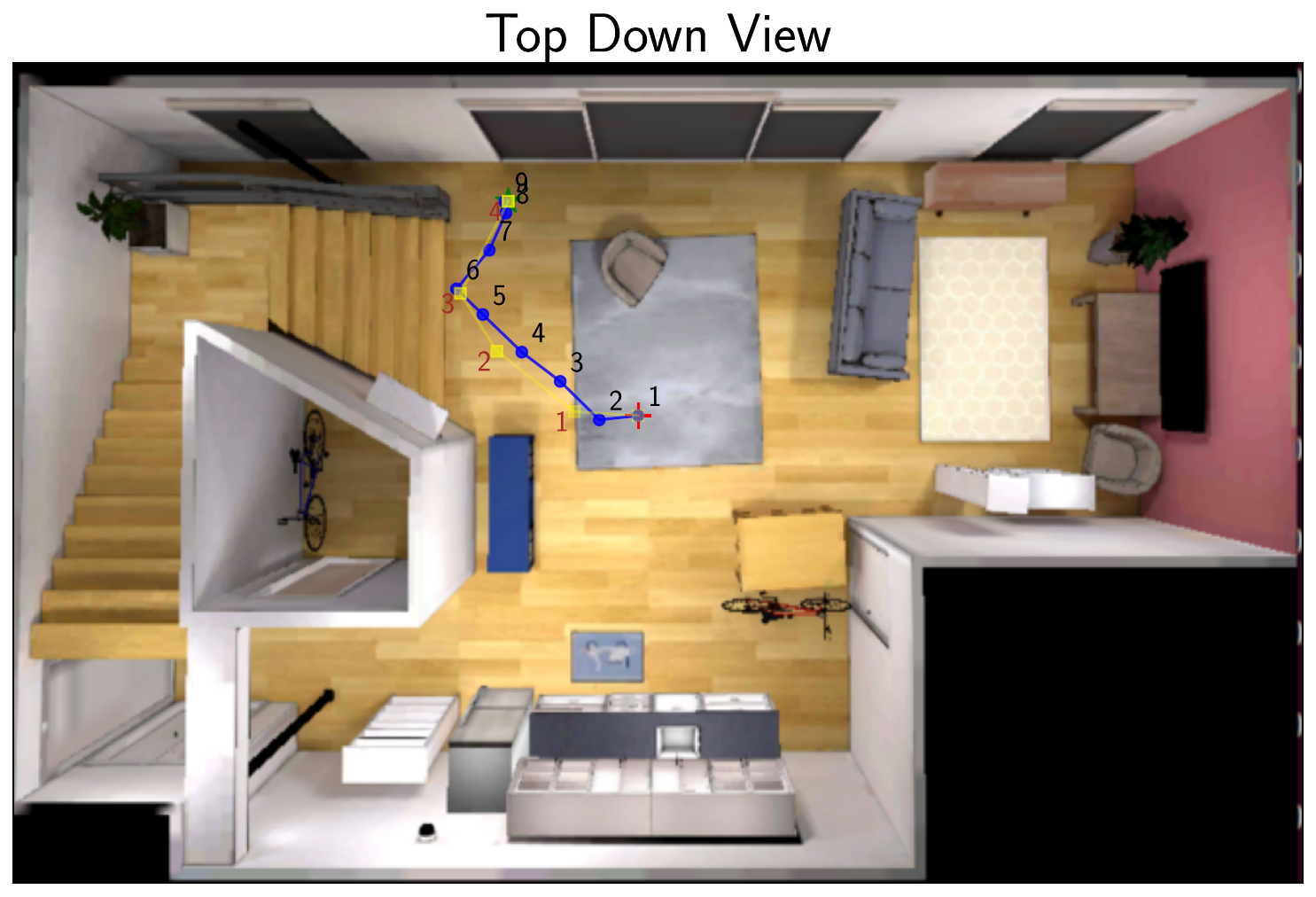}
       \label{fig:top_down_map_viz} 
    \end{subfigure}
    \begin{subfigure}[b]{0.48\textwidth}
       \includegraphics[width=\linewidth]{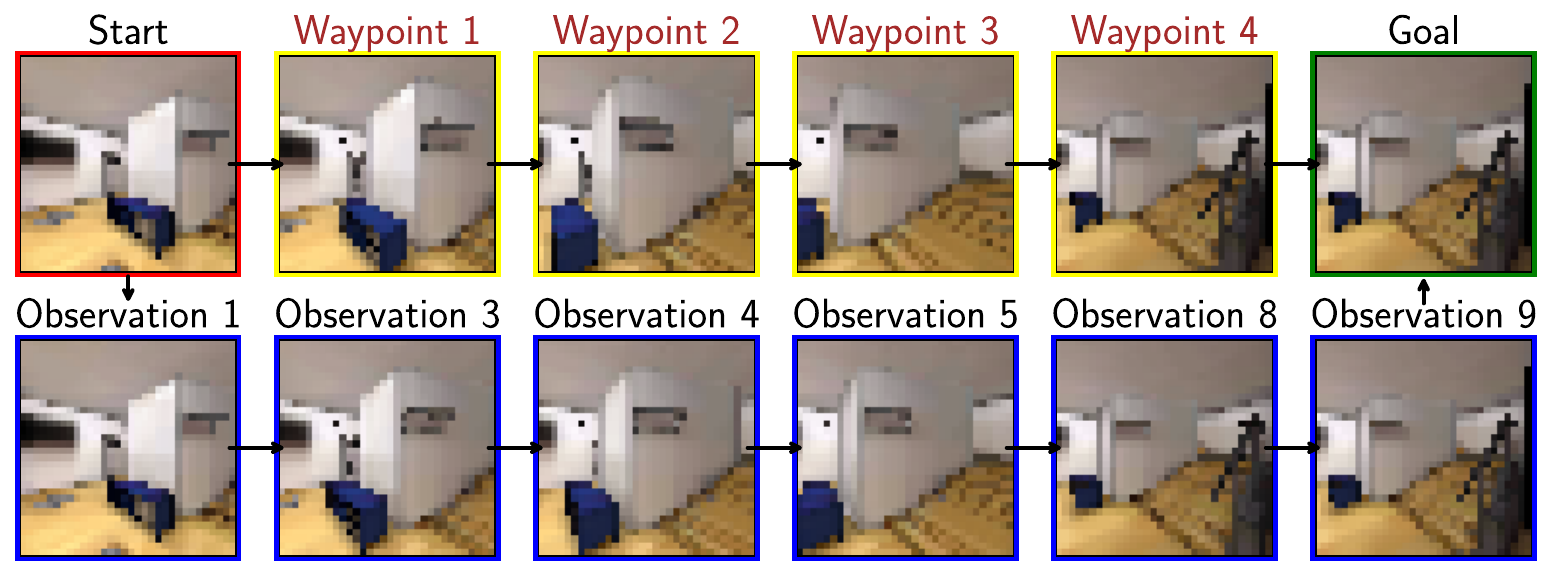}
       \label{fig:viz_nav}
    \end{subfigure}
    \caption{\textbf{Safe Visual Navigation}: The top-down map illustrates the agent's path and waypoints according to our approach. The first row displays the forward-facing start, goal, and suggested waypoint images. Bottom row shows the agent's forward-facing observations during execution. The waypoints help the agent avoid the couch and blue furniture. Note that the top-down view is not available to the agent.}
    \label{fig:visual_nav}
\end{figure}

\begin{table*}[ht]
\centering{}%
\begin{tabular}{cccccc@{}}
\toprule 
\multicolumn{2}{c}{\textbf{Problem Configurations}} & \multicolumn{4}{c}{\textbf{\begin{tabular}[c]{@{}c@{}}Methods\\ (Cumulative Cost (Success Rate))\end{tabular}}}\tabularnewline
\midrule 
\begin{tabular}{@{}c@{}}
Problem Type\tabularnewline
\end{tabular} & Agents & %
\begin{tabular}{@{}c@{}}
Unconstrained Policy\tabularnewline
\end{tabular} & %
\begin{tabular}{@{}c@{}}
Unconstrained Search (SoRB)\tabularnewline
\end{tabular} & %
\begin{tabular}{@{}c@{}}
Constrained Policy\tabularnewline
\end{tabular} & %
\begin{tabular}{@{}c@{}}
Constrained Search (Ours)\tabularnewline
\end{tabular}\tabularnewline
\midrule 
\multirow{4}{*}{Easy} & 1 & \textbf{0.46 $\pm$ 1.05 (100\%)} & 1.38 \textbf{$\pm$} 2.62 (100\%) & 0.47 \textbf{$\pm$} 1.07 (100\%) & 0.49 \textbf{$\pm$} 1.06 (100\%)\tabularnewline
 & 5 & N/A & 6.39 \textbf{$\pm$} 4.66 (100\%) & N/A & \textbf{1.68 \textbf{$\pm$} 1.21 (100\%)}\tabularnewline
 & 10 & N/A & 8.69 \textbf{$\pm$} 4.05 (100\%) & N/A & \textbf{2.24 \textbf{$\pm$} 1.03 (100\%)}\tabularnewline
 & 20 & N/A & 10.51 \textbf{$\pm$} 3.57 (100\%) & N/A & \textbf{2.72 \textbf{$\pm$} 0.80 (100\%)}\tabularnewline
\midrule 
\multirow{4}{*}{Medium} & 1 & 1.58 \textbf{$\pm$} 2.03 (100\%) & 2.30 \textbf{$\pm$} 3.02 (100\%) & \textbf{1.58 $\pm$ 2.01 (100\%)} & 1.59 \textbf{$\pm$} 1.99 (100\%)\tabularnewline
 & 5 & N/A & 5.12 \textbf{$\pm$} 3.65 (100\%) & N/A & \textbf{3.09 $\pm$ 1.50 (100\%)}\tabularnewline
 & 10 & N/A & 7.27 \textbf{$\pm$} 3.66 (98\%) & N/A & \textbf{4.11 $\pm$ 1.48 (98\%)}\tabularnewline
 & 20 & N/A & 8.73 \textbf{$\pm$} 3.73 (98\%) & N/A & \textbf{4.78 $\pm$ 1.33 (98\%)} \tabularnewline
\midrule 
\multirow{4}{*}{Hard} & 1 & 3.98 \textbf{$\pm$} 4.40 (100\%) & 4.05 \textbf{$\pm$} 4.19 (100\%) & 4.19 \textbf{$\pm$} 4.23 (100\%) & \textbf{3.09 $\pm$ 3.87 (100\%)}\tabularnewline
 & 5 & N/A & 8.58 \textbf{$\pm$} 3.79 (100\%) & N/A & \textbf{6.01 $\pm$ 1.96 (100\%)}\tabularnewline
 & 10 & N/A & 10.77 \textbf{$\pm$} 3.51 (100\%) & N/A & \textbf{7.24 $\pm$ 1.52 (100\%)}\tabularnewline
 & 20 & N/A & 11.96 \textbf{$\pm$} 3.31 (100\%) & N/A & \textbf{8.36 $\pm$ 1.58 (100\%)}\tabularnewline
\bottomrule
\end{tabular}
\caption{\textbf{2D Navigation}: Comparison of accumulated cost across different approaches with varying problem difficulty and number of agents. Our approach is consistently able to plan successful and safest paths for multiple agents. Note that we extended SoRB by integrating ith with CBS to plan waypoints for multiple agents. Lower cumulative costs and higher success rates are better. Not applicable to multi-agent policies without search, since MARL approaches perform poorly.}
\label{tab:central_obstacle}
\end{table*}

\begin{table*}[ht]
\begin{tabular}{@{}ccccccc@{}}
\toprule
\multicolumn{3}{c}{\textbf{Problem Configurations}} & \multicolumn{4}{c}{\textbf{\begin{tabular}[c]{@{}c@{}}Methods\\ (Cumulative Cost (Success Rate))\end{tabular}}} \\ \midrule
Map & \begin{tabular}[c]{@{}c@{}}Problem\\ Type\end{tabular} & Agents & Unconstrained Policy & Unconstrained  Search (SoRB) & Constrained Policy & Constrained Search (Ours) \\ \midrule
\multirow{9}{*}{\begin{tabular}[c]{@{}c@{}}SC2 \\ Staging 08\end{tabular}} & \multirow{3}{*}{Easy} & 1 & 3.16 $\pm$ 9.40 (100\%) & 2.14 $\pm$ 2.22 (98\%) & 1.85 $\pm$ 1.80 (100\%) & \textbf{1.28 $\pm$ 1.69 (100\%)} \\
 &  & 5 & N/A & 4.32 $\pm$ 2.29 (98\%) & N/A & \textbf{3.35 $\pm$ 1.48 (100\%)} \\
 &  & 10 & N/A & 5.36 $\pm$ 2.60 (96\%) & N/A & \textbf{4.02 $\pm$ 1.15 (100\%)} \\ \cmidrule(l){2-7} 
 & \multirow{3}{*}{Medium} & 1 & 4.98 $\pm$ 3.46 (98\%) & 5.15 $\pm$ 3.32 (98\%) & 4.77 $\pm$ 3.25 (100\%) & \textbf{3.35 $\pm$ 3.23 (100\%)} \\
 &  & 5 & N/A & 9.74 $\pm$ 11.13 (90\%) & N/A & \textbf{6.18 $\pm$ 1.64 (98\%)} \\
 &  & 10 & N/A & 14.23 $\pm$ 23.53 (84\%) & N/A & \textbf{7.30 $\pm$ 1.69 (98\%)} \\ \cmidrule(l){2-7} 
 & \multirow{3}{*}{Hard} & 1 & 13.41 $\pm$ 4.00 (96\%) & 12.35 $\pm$ 5.10 (90\%) & 10.43 $\pm$ 3.07 (100\%) & \textbf{9.62 $\pm$ 3.66 (100\%)} \\
 &  & 5 & N/A & 21.23 $\pm$ 14.64 (84\%) & N/A & \textbf{13.02 $\pm$ 2.38 (98\%)} \\
 &  & 10 & N/A & 27.37 $\pm$ 19.05 (62\%) & N/A & \textbf{13.87 $\pm$ 2.17 (92\%)} \\ \midrule
\multirow{9}{*}{\begin{tabular}[c]{@{}c@{}}SC3 \\ Staging 05\end{tabular}} & \multirow{3}{*}{Easy} & 1 & 1.87 $\pm$ 1.65 (98\%) & 2.23 $\pm$ 1.85 (98\%) & 1.80 $\pm$ 1.58 (100\%) & \textbf{1.25 $\pm$ 1.60 (100\%)} \\
 &  & 5 & N/A & 3.71 $\pm$ 1.96 (98\%) & N/A & \textbf{2.76 $\pm$ 1.36 (100\%)} \\
 &  & 10 & N/A & 4.20 $\pm$ 1.73 (98\%) & N/A & \textbf{3.32 $\pm$ 1.11 (100\%)} \\ \cmidrule(l){2-7} 
 & \multirow{3}{*}{Medium} & 1 & 5.06 $\pm$ 3.27 (100\%) & 4.87 $\pm$ 3.15 (100\%) & 4.38 $\pm$ 2.91 (100\%) & \textbf{2.66 $\pm$ 2.32 (100\%)} \\
 &  & 5 & N/A & 7.58 $\pm$ 2.16 (98\%) & N/A & \textbf{5.88 $\pm$ 1.38 (100\%)} \\
 &  & 10 & N/A & 8.83 $\pm$ 1.75 (98\%) & N/A & \textbf{7.28 $\pm$ 1.53 (96\%)} \\ \cmidrule(l){2-7} 
 & \multirow{3}{*}{Hard} & 1 & 14.86 $\pm$ 4.9 (98\%) & 13.16 $\pm$ 4.24 (100\%) & 8.59 $\pm$ 2.21 (98\%) & \textbf{6.47 $\pm$ 3.30 (98\%)} \\
 &  & 5 & N/A & 17.79 $\pm$ 2.68 (96\%) & N/A & \textbf{15.10 $\pm$ 2.19 (96\%)} \\
 &  & 10 & N/A & 19.56 $\pm$ 2.66 (94\%) & N/A & \textbf{16.71 $\pm$ 2.28 (94\%)} \\ \midrule
\multirow{9}{*}{\begin{tabular}[c]{@{}c@{}}SC3 \\ Staging 11\end{tabular}} & \multirow{3}{*}{Easy} & 1 & 1.20 $\pm$ 1.56 (100\%) & 1.27 $\pm$ 1.35 (100\%) & 1.20 $\pm$ 1.55 (100\%) & \textbf{0.78 $\pm$ 1.12 (100\%)} \\
 &  & 5 & N/A & 2.32 $\pm$ 1.11 (100\%) & N/A & \textbf{1.88 $\pm$ 0.90 (100\%)} \\
 &  & 10 & N/A & 3.00 $\pm$ 1.07 (100\%) & N/A & \textbf{2.54 $\pm$ 1.15 (100\%)} \\ \cmidrule(l){2-7} 
 & \multirow{3}{*}{Medium} & 1 & 3.66 $\pm$ 2.95 (98\%) & 4.19 $\pm$ 3.23 (98\%) & 3.40 $\pm$ 2.83 (98\%) & \textbf{1.95 $\pm$ 2.24 (100\%)} \\
 &  & 5 & N/A & 8.33 $\pm$ 2.37 (98\%) & N/A & \textbf{5.93 $\pm$ 2.40 (96\%)} \\
 &  & 10 & N/A & 10.23 $\pm$ 2.27 (96\%) & N/A & \textbf{8.13 $\pm$ 2.18 (94\%)} \\ \cmidrule(l){2-7} 
 & \multirow{3}{*}{Hard} & 1 & 16.41 $\pm$ 6.50 (100\%) & 14.75 $\pm$ 5.43 (100\%) & 11.59 $\pm$ 3.60 (100\%) & \textbf{10.93 $\pm$ 3.94 (100\%)} \\
 &  & 5 & N/A & 18.67 $\pm$ 2.51 (96\%) & N/A & \textbf{15.78 $\pm$ 2.16 (98\%)} \\
 &  & 10 & N/A & 20.37 $\pm$ 2.41 (92\%) & N/A & \textbf{16.94 $\pm$ 1.66 (92\%)} \\ \midrule
\multirow{9}{*}{\begin{tabular}[c]{@{}c@{}}SC3 \\ Staging 15\end{tabular}} & \multirow{3}{*}{Easy} & 1 & 1.90 $\pm$ 1.76 (100\%) & 2.08 $\pm$ 1.73 (100\%) & 1.87 $\pm$ 1.77 (100\%) & \textbf{1.47 $\pm$ 1.39 (100\%)} \\
 &  & 5 & N/A (100\%) & 3.82 $\pm$ 3.53 (100\%) & N/A & \textbf{2.77 $\pm$ 1.32 (100\%)} \\
 &  & 10 & N/A & 4.83 $\pm$ 3.68 (100\%) & N/A & \textbf{3.63 $\pm$ 1.39 (100\%)} \\ \cmidrule(l){2-7} 
 & \multirow{3}{*}{Medium} & 1 & 3.76 $\pm$ 3.42 (100\%) & 4.16 $\pm$ 2.79 (100\%) & 3.23 $\pm$ 2.51 (100\%) & \textbf{2.01 $\pm$ 1.75 (100\%)} \\
 &  & 5 & N/A & 6.70 $\pm$ 1.81 (100\%) & N/A & \textbf{5.18 $\pm$ 2.32 (100\%)} \\
 &  & 10 & N/A & 10.21 $\pm$ 14.89 (96\%) & N/A & \textbf{6.66 $\pm$ 1.59 (100\%)} \\ \cmidrule(l){2-7} 
 & \multirow{3}{*}{Hard} & 1 & 13.83 $\pm$ 4.35 (100\%) & 14.37 $\pm$ 11.06 (98\%) & 9.75 $\pm$ 2.74 (100\%) & \textbf{7.61 $\pm$ 3.10 (100\%)} \\
 &  & 5 & N/A & 19.23 $\pm$ 11.03 (88\%) & N/A & \textbf{11.63 $\pm$ 2.70 (100\%)} \\
 &  & 10 & N/A & 19.70 $\pm$ 3.73 (84\%) & N/A & \textbf{12.85 $\pm$ 2.12 (98\%)} \\ \bottomrule
\end{tabular}
\caption{\textbf{Safe Visual Navigation}: Comparison of accumulated cost across different approaches with varying problem difficulty and number of agents. Lower cumulative costs and higher success rates are better. Not applicable to multi-agent policies without search, since MARL approaches perform poorly.}
\label{tab:visual_nav}
\end{table*}

\section{Experiments}
\label{experiments}

Our experiments address the following questions to evaluate the effectiveness of our approach:

\begin{itemize}[noitemsep, align=left]
\item[\textbf{Q1:}] Can our approach demonstrate safer agent behavior compared to other methods?
\item[\textbf{Q2:}] Does our approach sacrifice its ability to reach distant goals when balancing the accumulated cost compared to other baselines?
\item[\textbf{Q3:}] Does the suggested approach effectively demonstrate superior performance when scaled to larger number of agents compared to other methods?
\end{itemize}

For the single-agent experiments, we compare our approach against a GCRL policy, SoRB and a safe GCRL policy. For the multi-agent experiments, we evaluated both off-policy and on-policy Multi-Agent RL (MARL) approaches, including MADDPG \cite{lowe2020multiagentactorcriticmixedcooperativecompetitive} and MAPPO \cite{yu2022surprisingeffectivenessppocooperative}. However, these methods exhibited the same limitations as GCRL approaches, ultimately failing to enable multiple agents to reach their goals. Hence, we compare our approach with an extended version of SoRB that utilises CBS to generate intermediate waypoints for multiple agents. Finally, we compare our approach with these baselines over two different tasks: a 2D navigation problem and a visual navigation problem that involves using images as observations.

\textbf{Environments}. We use a 2D navigation environment with Central Obstacle map, where the agent observations are their locations, represented as $s = (x, y) \in \mathbb{R}^2$. The agent's actions, $a = (dx, dy) \in [-1, 1]^2$, are added to the current position at each time step to update their location. For the visual navigation problems, we selected four distinct environments from the rich, interactive digital-twin ReplicaCAD dataset \cite{replica19arxiv}, integrated with Habitat-Sim \cite{szot2021habitat,habitat19iccv}. In these scenarios, the agent's observations consist of RGB panoramic images created by concatenating $32 \times 32$ first-person views from each of the four cardinal directions. The action space for the agent in the visual navigation problems is the same as in the 2D navigation scenario. For all experiments, we choose the cost limit $\delta=10$, and we use a cost function that linearly decreases from the obstacle boundaries.
\begin{equation}
    C(s) = 
        \begin{cases}
            2-2h/r & 0 \le h \le r \\
            0 & \textnormal{otherwise}
        \end{cases}
    \label{eq:cost_f}
\end{equation}
where $h: \mathcal{S} \rightarrow \mathbb{R}_{\ge0}$ is the distance to the boundary of the nearest obstacle and $r \in \mathbb{R}_{+}$ is the radius of influence of each obstacle. In our experiments, $r$ is set to 10 for 2D navigation and 1 for visual navigation tasks respectively.

Figure \ref{fig:pointenv_illustration_together} illustrates example trajectories generated by our approach and the baselines in single-agent and multi-agent settings in the 2D navigation task for a qualitative comparison. Figure \ref{fig:pointenv_illustration_together} shows that, while the baseline methods often fail to generate safer paths when the sampled start and goal locations fall within high-cost regions, our approach successfully identifies a path that, while counter-intuitive from the perspective of the RL agent, aligns with human intuition. By navigating around the high-cost regions, our approach effectively minimizes the accumulated cost.

We now examine the effectiveness of our approach on the more challenging visual navigation problem as shown in Figure \ref{fig:visual_nav}. Figures \ref{fig:habitatenv_multi_illustration} present qualitative comparison with example trajectories generated by both the baselines and our approach. The results are consistent with those from the 2D navigation problem, demonstrating that our approach enables agents to re-route their paths away from obstacles, such as furniture in the room, leading to safer navigation behavior.

Our quantitative experiments involved randomly generating problems with varying difficulty levels according to the distance and cost predictions from the trained value functions, placing start and goal positions closer together for easier tasks and further apart for harder ones to increase complexity. Each experiment consisted of 50 different problem instances for each difficulty level. In our multi-agent experiments, we compared our approach with an extended version of SoRB, varying the number of agents from 5 to 20 for the 2D navigation problem, and from 5 to 10 agents for the visual navigation problem.

On the simpler 2D navigation problem, Table \ref{tab:central_obstacle} shows that our approach consistently generates safer paths with the lowest accumulated cost compared to the other baselines, both in single-agent and multi-agent settings. This safer behavior becomes more pronounced as the number of agents increases, with no loss in success rate.

For the challenging visual navigation problems, Table \ref{tab:visual_nav} demonstrates that our approach again outperforms the baselines by generating safer paths for the agents in both single and multi-agent settings. Notably, we observed a significant drop in success rate for SoRB with an increased number of agents, particularly in the SC2 Staging 08 map. This is likely due to longer re-routing times as agents avoid each other, leading to violation of the time limits. However, as shown in Table \ref{tab:visual_nav}, our approach consistently maintains safer paths for the agents, with the safety margin improving as the number of agents increases.

Regarding success rate, our approach manages to maintain near-perfect scores, barring a slight drop in performance when scaling to larger numbers of agents. This behavior is expected, as with more agents, the traversable regions become narrower, making it more challenging for CBS to find collision-free paths. Consequently, agents may need to wait longer at some assigned waypoints, resulting in increased steps to reach their goals and a higher likelihood of exceeding time limits. Based on these results, we conclude that our approach not only plans safer paths for agents but does so with minimal loss in their ability to reach their goals. It also demonstrates consistent trends when scaling to larger numbers of agents, effectively addressing \textbf{Q1}, \textbf{Q2}, and \textbf{Q3}.

\section{Conclusions}
In conclusion, this work introduces a novel hierarchical framework that combines high-level multi-agent planning using MAPF approaches with low-level control through safe GCRL policies. The framework leverages learned distance and cost critics to automatically generate and annotate dense graph networks from the replay buffer using self-sampling and training. At the high level, our approach utilizes CBS to plan efficient and safe waypoints for multiple agents, while at the low level, the safe GCRL policy guides the agents to follow these waypoints toward distant goal-reaching tasks safely. This bi-level safety control ensures safer multi-agent behavior throughout execution, effectively tackling scenarios ranging from simple 2D navigation to more complex visual navigation challenges.

\addtolength{\textheight}{-10cm}

\bibliographystyle{./ICRA_2024/IEEETransactions}
\bibliography{IEEEReferences}

\end{document}

\section{Appendix}

\subsection{Hyperparameters}

\begin{table*}[ht]
  \centering
  \caption{Hyperparameters and Training Settings for Visual Navigation}
  \label{tab:hyperparameters}
  \begin{tabular}{ll}
    \toprule
    \textbf{Parameter} & \textbf{Value} \\
    \midrule
    Actor Learning Rate           & 1e-5 \\
    Actor Update Interval         & 1 \\
    Critic Learning Rate          & 1e-4 \\
    Cost Critic Learning Rate     & 1e-4 \\
    Distance Critic Bins          & 20 \\
    Cost Critic Bins              & 40 \\
    Targets Update Interval       & 5 \\
    Polyak Update Coefficient     & 0.05 \\
    Initial Lagrange Multiplier   & 0 \\
    Lagrange Learning Rate        & 0.035 \\
    Optimizer                     & Adam \\
    Visual Inputs Dimensions      & (4, 32, 32, 4) \\
    Replay Buffer Size            & 100000 \\
    Batch Size                    & 64 \\
    Initial Collect Steps         & 1000 \\
    Training Iterations           & 600000 \\
    Neural Network Architecture   & Conv(16, 8, 4) + Conv(32, 4, 4) + FC(256) \\
    Maximum Episode Steps         & 20 \\
    \bottomrule
  \end{tabular}
\end{table*}

\subsection{Limitations}

One key limitation of our approach is the absence of strict risk-bounded guarantees on the cumulative risk incurred by all agents. In real-world scenarios, an ideal system would accept user input regarding risk averseness and automatically adjust its behavior accordingly. Although there are existing methods for multi-objective optimization, to our knowledge, none guarantee bounded execution risk while also maximizing reward. Additionally, the constrained low-level policy was trained on a single agent in a fixed environment, so its applicability in dynamic settings has not been empirically validated—even though SoRB has demonstrated effectiveness in diverse environments. In fast-changing, dynamic settings, the safer behaviors provided by our approach may be less effective.

\subsection{Real-World Connections}

To validate the approach in real-world settings, it is essential to ensure that agents adhere to the timings of each waypoint determined by the high-level CBS search in order to avoid collisions with other agents. If deviations from the nominal trajectories occur, low-level agents should utilize individual collision avoidance strategies by communicating their positions with one another, ensuring that when agents come too close, one can yield until the other has passed. Moreover, for smoother deployment, the operating height of the agents must be considered, as it defines the obstacle boundaries relevant to the safety function. Incorporating the height (z-coordinate) could lead to more natural plans, such as enabling drones to avoid obstacles by flying over them.